\documentclass[twocolumn]{article}
\usepackage{graphicx}
\usepackage{subfigure}
\usepackage{natbib}
\usepackage{microtype}
\usepackage{booktabs}
\usepackage{siunitx}
\usepackage{hyperref}
\usepackage{mathtools}
\usepackage{amsmath}
\usepackage{amsthm}
\usepackage{amssymb}
\usepackage{cleveref}
\usepackage{cancel}
\usepackage{color}
\usepackage{tabularx}
\usepackage{tikz}
\usepackage{rotating}
\usepackage{xfrac}

\begingroup
    \makeatletter
    \@for\theoremstyle:=theorem,corollary,lemma,model,definition,remark,plain\do{%
        \expandafter\g@addto@macro\csname th@\theoremstyle\endcsname{%
            \setlength\thm@preskip\parskip
            \setlength\thm@postskip\parskip
            \addtolength\thm@preskip\parskip
            }%
        }
\endgroup

\newcommand{\R}{\mathbb{R}}   

\newcommand{\e}{\varepsilon}   
\renewcommand{\O}{\mathcal{O}} 

\renewcommand{\max}{\operatorname{max}}
\newcommand{\atan}{\operatorname{atan}}

\newcommand{\tr}{\operatorname{tr}}

\newcommand{\KL}{\operatorname{D}_{\text{KL}}}

\renewcommand{\ss}[1]{_\text{#1}}
\newcommand{\sd}[1]{\, \mathrm{d} #1}

\newcommand{\cond}{\, | \,}
\renewcommand{\ll}{\left}
\newcommand{\rr}{\right}
\newcommand{\la}{\langle}
\newcommand{\ra}{\rangle}
\newcommand{\phan}[1]{\hphantom{#1\;}}
\newcommand{\T}{\text{\textsf{T}}}

\newtheorem{model}{Model}
\crefname{model}{Model}{Models}
\Crefname{model}{Model}{Models}
\crefname{equation}{Equation}{Equations}
\Crefname{equation}{Equation}{Equations}
\crefname{figure}{Figure}{Figures}
\Crefname{figure}{Figure}{Figures}
\crefname{table}{Table}{Tables}
\Crefname{table}{Table}{Tables}
\crefname{section}{Section}{Sections}
\Crefname{section}{Section}{Sections}
\crefname{appendix}{Appendix}{Appendices}
\Crefname{appendix}{Appendix}{Appnedices}

\usepackage{parskip}
\newif\ifpdf
\ifx\pdfoutput\undefined
\else
   \ifx\pdfoutput\relax
   \else
     \ifcase\pdfoutput
     \else
       \pdftrue
     \fi
   \fi
\fi
\ifpdf
\fi
\makeatletter

\def\section{\@startsection{section}{1}{\z@}{-0.12in}{0.02in}
             {\large\bf\raggedright}}
\def\subsection{\@startsection{subsection}{2}{\z@}{-0.10in}{0.01in}
                {\normalsize\bf\raggedright}}
\def\subsubsection{\@startsection{subsubsection}{3}{\z@}{-0.08in}{0.01in}
                {\normalsize\sc\raggedright}}
\def\paragraph{\@startsection{paragraph}{4}{\z@}{1.5ex plus
  0.5ex minus .2ex}{-1em}{\normalsize\bf}}
\def\subparagraph{\@startsection{subparagraph}{5}{\z@}{1.5ex plus
  0.5ex minus .2ex}{-1em}{\normalsize\bf}}
\skip\footins 9pt
\def\footnoterule{\kern-3pt \hrule width 0.8in \kern 2.6pt }
\labelwidth\leftmargini\advance\labelwidth-\labelsep \labelsep 5pt
\def\@listi{\leftmargin\leftmargini}
\def\@listii{\leftmargin\leftmarginii
   \labelwidth\leftmarginii\advance\labelwidth-\labelsep
   \topsep 2pt plus 1pt minus 0.5pt
   \parsep 1pt plus 0.5pt minus 0.5pt
   \itemsep \parsep}
\def\@listiii{\leftmargin\leftmarginiii
    \labelwidth\leftmarginiii\advance\labelwidth-\labelsep
    \topsep 1pt plus 0.5pt minus 0.5pt
    \parsep \z@ \partopsep 0.5pt plus 0pt minus 0.5pt
    \itemsep \topsep}
\def\@listiv{\leftmargin\leftmarginiv
     \labelwidth\leftmarginiv\advance\labelwidth-\labelsep}
\def\@listv{\leftmargin\leftmarginv
     \labelwidth\leftmarginv\advance\labelwidth-\labelsep}
\def\@listvi{\leftmargin\leftmarginvi
     \labelwidth\leftmarginvi\advance\labelwidth-\labelsep}
\belowdisplayskip \abovedisplayskip
\def\@normalsize{\@setsize\normalsize{11pt}\xpt\@xpt}
\def\small{\@setsize\small{10pt}\ixpt\@ixpt}
\def\footnotesize{\@setsize\footnotesize{10pt}\ixpt\@ixpt}
\def\scriptsize{\@setsize\scriptsize{8pt}\viipt\@viipt}
\def\tiny{\@setsize\tiny{7pt}\vipt\@vipt}
\def\large{\@setsize\large{14pt}\xiipt\@xiipt}
\def\Large{\@setsize\Large{16pt}\xivpt\@xivpt}
\def\LARGE{\@setsize\LARGE{20pt}\xviipt\@xviipt}
\def\huge{\@setsize\huge{23pt}\xxpt\@xxpt}
\def\Huge{\@setsize\Huge{28pt}\xxvpt\@xxvpt}
\makeatother
\usepackage{import}
\usepackage{standalone}

\begin{document}
\setcitestyle{authoryear,round,citesep={;},aysep={,},yysep={;}}

\title{Learning Causally-Generated Stationary Time Series}
\author{
    Wessel Bruinsma%
    \thanks{Research primarily conducted whilst at Invenia Labs, Cambridge, UK.}\\
    University of Cambridge \\
    \texttt{wpb23@cam.ac.uk} \\
    \and
    Richard E. Turner\\
    University of Cambridge \\
    \texttt{ret26@cam.ac.uk}
}
\maketitle


\begin{abstract}
    We present the Causal Gaussian Process Convolution Model (CGPCM), a doubly nonparametric model for causal, spectrally complex dynamical phenomena. The CGPCM is a generative model in which white noise is passed through a causal, nonparametric-window moving-average filter, a construction that we show to be equivalent to a Gaussian process with a nonparametric kernel that is biased towards causally-generated signals. We develop enhanced variational inference and learning schemes for the CGPCM and its previous acausal variant, the GPCM \citep{Tobar:2015:Learning_Stationary}, that significantly improve statistical accuracy. These modelling and inferential contributions are demonstrated on a range of synthetic and real-world signals.
\end{abstract}

\section{Introduction}
One of the major goals of statistical inference is to develop models of dynamical phenomena that can be used for prediction and system identification. Two key characteristics of the physical systems underlying natural and manmade dynamical phenomena are that they are \textit{causal} and \textit{spectrally complex}. Causal systems, where at any point in time the output of the system can only depend on past values of the input, are the only type that are physically realisable. Spectrally complex systems, which have rich power spectral densities, arise because many physical systems have numerous resonances spanning many time scales. When developing statistical models for such phenomena, this prior knowledge---causality and spectral richness---should be leveraged in order to exclude any unrealisable system from the model prior, whilst at the same time allowing the model to have the capacity to capture varied spectral content that may be only slowly revealed as more data are seen. The goal of this paper is to develop Gaussian process (GP) models together with associated inference and learning schemes that serve this purpose.

Gaussian processes are a widely-used model for stationary time series. They place a prior distribution over the latent function underlying the time series $f:\R \to \R$ by assuming that any finite collection of function values $f(t_1),\ldots,f(t_n)$ is multivariate Gaussian distributed.
GPs are nonparametric models, which means that as opposed to parametric models, there is no finite number of parameters that parametrises a GP.
Instead, the number of parameters grows with the amount of evidence that is accumulated. This property allows GPs to learn complex functions if plenty of evidence is available, and conversely makes them robust against overfitting if only little evidence is at hand.
The key modelling decision in using a Gaussian process is the choice of covariance $k_f(t,t')$ between any two function values $f(t)$ and $f(t')$; $k_f$ is commonly called the \textit{kernel} of $f$.
The kernel encodes prior information about the function $f$ into the model. Consequently, a large research effort has been devoted towards developing flexible kernels \citep{Duvenaud:2014:Automatic_Construction,Wilson:2013:Spectral_Mixture,Tobar:2015:Inter-Domain_Inducing,Tobar:2015:Learning_Stationary}.

Of particular interest for the current application are GP models that additionally use nonparametric models for the power spectra of the GPs---doubly nonparametric models---as they have the capacity to flexibly model signals with arbitrary spectral complexity. One approach employs a Dirichlet process mixture model for the power spectra \citep{Oliva:2015:Bayesian_Nonparametric_Kernel-Learning}, but it is not clear whether it is possible to build the causality constraint into such a construction. An alternative approach induces a nonparametric power spectrum by placing a Gaussian process prior over a linear system's impulse response function \citep{Tobar:2015:Learning_Stationary}. Critically, this alternative approach, called the Gaussian Process Convolution Model (GPCM), does not incorporate a causality constraint. In this paper we revisit the GPCM and explicitly build in a causality constraint, whilst retaining the flexibility of modelling the kernel nonparametrically. In addition to these modelling contributions, we develop enhanced variational inference and learning schemes---collapsed variational bounds and structured approximating distributions---for the causal and acausal variants of the GPCM that significantly improve statistical accuracy. These modelling and inferential contributions are demonstrated on a range of synthetic and real-world signals.

\section{Modelling Causally-Generated Stationary Time Series}
Consider the problem of modelling a stationary time series $f$. Motivated by the fact that many dynamical systems in nature can accurately be described by an initial value problem or a linear system, we let $f$ be the solution to a time-invariant linear initial value problem with causal Green's function $h$ and forcing function $x$, or equivalently the system response of a time-invariant linear system with causal impulse response $h$ and excitation $x$:
\begin{align} \label{eq:model}
    f(t) = \int^t h(t- \tau)x(\tau)\sd{\tau}.
\end{align}
For parsimony, we denote integration from negative infinity and to positive infinity by omitting the respective limit from the integral throughout.
Note that $f(t)$ depends on $x(\tau)$ only for $\tau \le t$; this reveals \cref{eq:model}'s causal nature.

In this paper we consider the case that $x$ is white noise; that is, informally denoted, $x \sim \mathcal{GP}(0,\delta(t-t'))$ where $\delta$ denotes the Dirac delta function. In that case \cref{eq:model} can be interpreted as a Gaussian process with zero mean function and stationary covariance function
\begin{align}
    k_{f\cond h}(t,t')&= \int^t \!\!\!\! \int^{t'}\!h(t- \tau) \vspace{-1mm} h(t'- \tau')
        \mathbb{E}[ x(\tau) x(\tau')]\sd{\tau'} \sd{\tau} \nonumber \\
    &= \int^{t \land t'} \! h(t - \tau) h(t' - \tau) \sd{\tau} \nonumber \\
    &= \int_0 h(|t - t'| + \tau)h(\tau) \sd{\tau} \nonumber \\
    &= k_{f\cond h}(t-t')  \label{eq:kernel}
\end{align}
where $t \land t'$ denotes the minimum of $t$ and $t'$. The restriction of $x$ to white noise is without loss of generality if $h$ is sufficiently rich, since any causal Gaussian process can be represented in this form. An alternative lens through which to view the model is as the continuous-time generalisation of a causal moving-average filter.

One of the focusses in this work is to develop flexible prior distributions over the filter $h$ that can support essentially arbitrarily complex structure. We therefore choose to model it using a Gaussian process $h \sim \mathcal{GP}(0,k_h(t,t'))$. The covariance function $k_h(t,t')$ describing the prior over the filter should be carefully chosen. One important constraint on the covariance function comes from the fact that every real-world signal $f$ has finite power. This can be satisfied by letting the filter decay to zero at infinity sufficiently quickly \citep{Tobar:2015:Learning_Stationary}: let
\begin{align*}
    g &\sim \mathcal{GP}(0,k_g(t-t')),&
    h\cond g &= w g
\end{align*}
where $w(t)= \exp({- \alpha t^2})$, so that the filter is restricted to a Gaussian window. Then
\begin{align*}
    \mathbb{V}[f(t)]
    &= \int_{0}\exp({-2 \alpha t^2})\mathbb{V}[g(\tau)]\sd{\tau} \\
    &\le k_g(0) \int_0 \exp({- 2 \alpha t^2}) \sd{\tau}
    < \infty,
\end{align*}
which could be infinite otherwise.
Equivalently, we let
\begin{align*}
     k_h(t,t')&=\exp({- \alpha (t^2 + t^{\prime 2})})k_g(t-t').
\end{align*}
To retain flexibility in the prior on $h$, we let $k_g$ be an exponentiated quadratic. We thus have that
\begin{align*}
     k_h(t,t')&=\exp({- \alpha (t^2 + t^{\prime 2}) - \gamma(t-t')^2}).
\end{align*}
Note that $\alpha$ determines the typical temporal extent of the filter and $\gamma$ the typical time-scale over which it varies. Further note that the assumption that $h$ decays to zero at infinity also serves to make inference well posed: if $h$ were not to decay to zero at infinity, any shifted version of the filter would result in an identical statistical model for the data, meaning that the model is unidentifiable.

The prior distributions over $x$ and $h$ induce a prior distribution on $f$. Further including a scale $\sigma_f$ to control $f$'s prior power, we call this prior on $f$ the Causal Gaussian Process Convolution Model (CGPCM). To recapitulate, the CGPCM admits the following two equivalent formulations:

\begin{model}[CGPCM (First Formulation)] \label{mod:cgpcm}
    \begin{gather*}
        x \sim \mathcal{GP}(0,\delta(t-t')), \quad
        h \sim \mathcal{GP}(0, k_h(t,t')), \\
        f\cond h, x = \sigma_f \int^t h(t- \tau)x(\tau)\sd{\tau}.
    \end{gather*}
\end{model}
\begin{model}[CGPCM (Second Formulation)] \label{mod:cgpcm2}
    \begin{align*}
        h &\sim \mathcal{GP}(0, k_h(t,t')), \\
        f \cond h &\sim \mathcal{GP}\left(0,  \sigma_f^2\int_0 h(|t-t'|+\tau)h(\tau)\sd{\tau} \right).
    \end{align*}
\end{model}

\begin{figure}[t]
    \includegraphics[width=\linewidth]{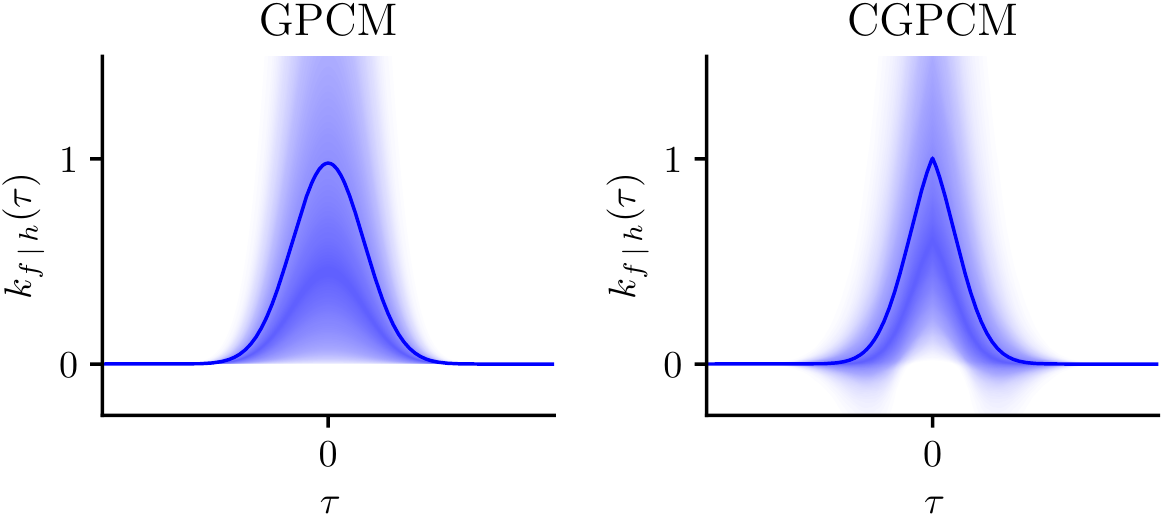}
    \caption{Prior distribution over kernels in the GPCM and CGPCM. Lines correspond to means and gradients indicate marginal variance.}
    \label{fig:prior_kernel}
\end{figure}
\begin{figure}[t]
    \includegraphics[width=\linewidth]{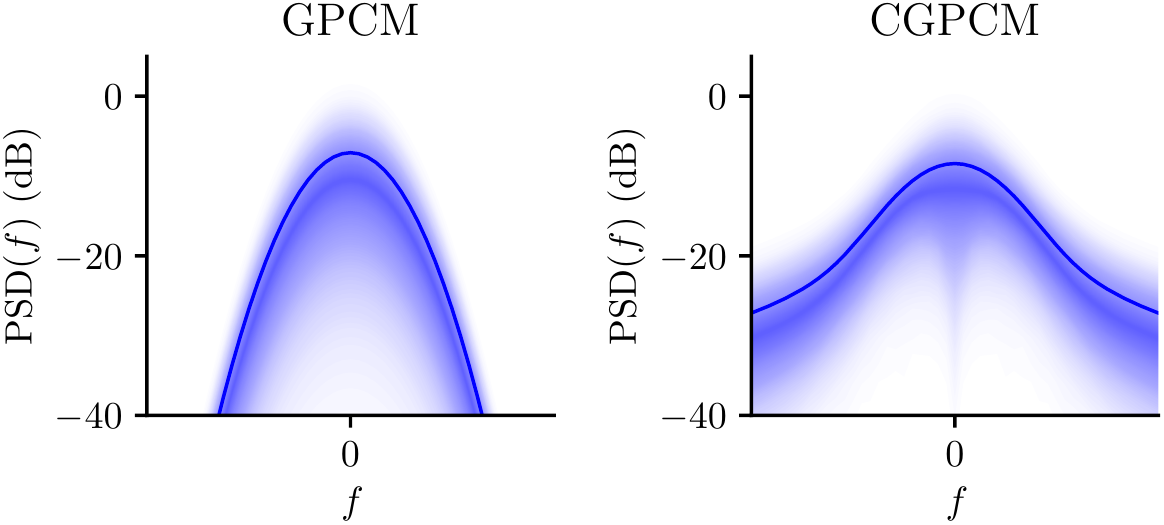}
    \caption{Prior distribution over PSDs in the GPCM and CGPCM. Lines correspond to means and gradients indicate marginal variance.}
    \label{fig:prior_psd}
\end{figure}

\Cref{mod:cgpcm2} reveals the CGPCM as a Gaussian process in which the kernel, or equivalently the power spectral density (PSD), is modelled nonparametrically. \Cref{fig:interpolation} illustrates the generative process of the CGPCM: First, a filter $h$ is generated. Then, the kernel $k_{f\cond h}$ is constructed. Finally, a sample $f\cond h$ is drawn from $\mathcal{GP}(0,k_{f \cond h}(t-t'))$. For both the GPCM and CGPCM, \cref{fig:prior_kernel} visualises the prior over kernels, and \cref{fig:prior_psd} visualises the prior over PSDs. Note that for the GPCM, the prior over PSDs confidently shows a low-pass structure, whereas that for the CGPCM shows support for more slowly decaying spectra. We will return to this observation in \cref{subsec:roughness}.

The CPGCM is similar to the latent force model presented by \citet{Alvarez:2009:Latent_Force_Models}; however, whereas we let $h$ be free form and let $x$ be white noise, \citet{Alvarez:2009:Latent_Force_Models} specify $h$ deterministically and let $x$ be free form.

\subsection{Sampling from the CGPCM}
Sampling from the CGPCM is challenging, because the integrals in \cref{mod:cgpcm,mod:cgpcm2} depend on the entirety of $h$ and $x$. To resolve this issue, we follow \citet{Tobar:2015:Learning_Stationary} and approximate \cref{eq:kernel} using Bayesian quadrature \citep{Minka:2000:Quadrature_GP} by conditioning on finitely many values
\begin{align*}
    u=(h(t_{u,1}),\ldots,h(t_{u,n_u}))\sim \mathcal{N}(0,K_u),
\end{align*}
also called \textit{inducing points} \citep{Titsias:2009:Variational_Learning} for $h$, which induce a distribution over covariance functions. The mean of this distribution is an accurate approximation for $k_{f\cond h}$ if sufficiently many inducing points are used:
\begin{align*}
    k_{f\cond h}(r)
    &\approx \mathbb{E}[k_{f\cond h}(r)\cond u]
    = \int_{0} \mathbb{E}[h(|r| + \tau) h(\tau)\cond u] \sd{\tau} \\
    &= \int_0 k_h(|r| + \tau, \tau) \sd{\tau} \\
    &\phan{=}+ \mathrm{trace} \left( M^u \int_{0} k_h(t_u, |r|+ \tau)k_h(\tau, t_u^\T) \sd{\tau} \right)
\end{align*}
where $M^{u}=K_u^{-1}uu^\T K_u^{-1}-K_u$. This expression was used to compute the model samples shown in \cref{fig:prior_psd,fig:interpolation,fig:prior_kernel}.

\subsection{Roughness of Sample Paths}
\label{subsec:roughness}
In this section we show that the CGPCM can capture both differentiable and nondifferentiable phenomena, the latter in various ``levels of roughness''. In this regard, the model is more flexible than the original GPCM, whose sample paths are almost surely differentiable. Intuitively, due to the convolution in \cref{eq:model} being causal, the white noise $x(t)$ entering the filter $h$ at $t$ is not smoothed out and can have a large effect if $|h(0)|>0$. Note that the fact that only the CGPCM can capture nondifferentiable phenomena is consistent with \cref{fig:prior_psd}: only the CGPCM has support for power at higher frequencies.

Let $h$ be a fixed filter that decays to zero at infinity. We claim that $f$'s sample paths are almost surely everywhere differentiable if $h(0)=0$, and almost surely nowhere differentiable if $h(0)\neq 0$. To show this, let
\begin{align*}
    g(r) = \int_0 h(r + \tau) h(\tau) \sd{\tau}
\end{align*}
so that
\begin{align*}
    k_{f\cond h}(r)
    &= g(|r|) \\
    &= g(0) + g'(0)|r| + \frac{1}{2}g''(0)r^2 + \O(|r|^3).
\end{align*}
Then, according to Theorem 2.6 and Example 2.3 in Section 2.3.1.2 by \citet{Lindgren:2006:Lectures_on_Stationary_Stochastic_Processes}, $f$'s sample paths are almost surely everywhere differentiable if $g'(0)=0$; otherwise, $f$ is not even mean square differentiable, in which case $f$'s sample paths are almost surely nowhere differentiable, according to Theorem 5 in \citep{Cambanis:1973:On_Some_Continuity_and_Differentiability}. To show the claim, integration by parts yields that
\begin{align*}
    g'(0)&= \int_0 h'(\tau) h(\tau) \sd{\tau}
    = \ll[ h^2(\tau)\rr]_0 - \int_0 h(\tau) h'(\tau) \sd{\tau} \\
    &= \lim_{\tau \to \infty}h^2(\tau) -h^2(0) - g'(0)
    = -h^2(0) - g'(0)
\end{align*}
where we used the fact that $h$ decays to zero at infinity.
Therefore, $g'(0) = -\frac{1}{2} h^2(0)$, so the claim is shown.

In the case that $h(0)\neq 0$, we can locally approximate $f$ by a $\sigma$-scaled Wiener process; this scale $\sigma$ then provides a quantification of the roughness of the sample paths. Specifically, given that the variance of an increment of a Wiener process is equal to the increment's length, we compute that
\begin{align*}
    \sigma^2
    &= \lim_{\e \downarrow 0} \frac{1}{\e}\mathbb{V}[f(t+\e)-f(t)] \\
    &= \lim_{\e \downarrow 0} \frac{2}{\e}(k_{f\cond h}(0) - k_{f\cond h}(\e))
    = -2g'(0)
    = h^2(0).
\end{align*}

In summary, the CGPCM models differentiable phenomena if $h(0)=0$ and nondifferentiable phenomena if $h(0)\neq 0$. In the latter case, $|h(0)|$ quantifies the level of roughness. Performing inference for the filter will therefore automatically infer the roughness of the underlying process from possibly noisy data.

\begin{figure*}[t]
    \vskip 0.1in
    \centering
    \includegraphics[width=\linewidth]{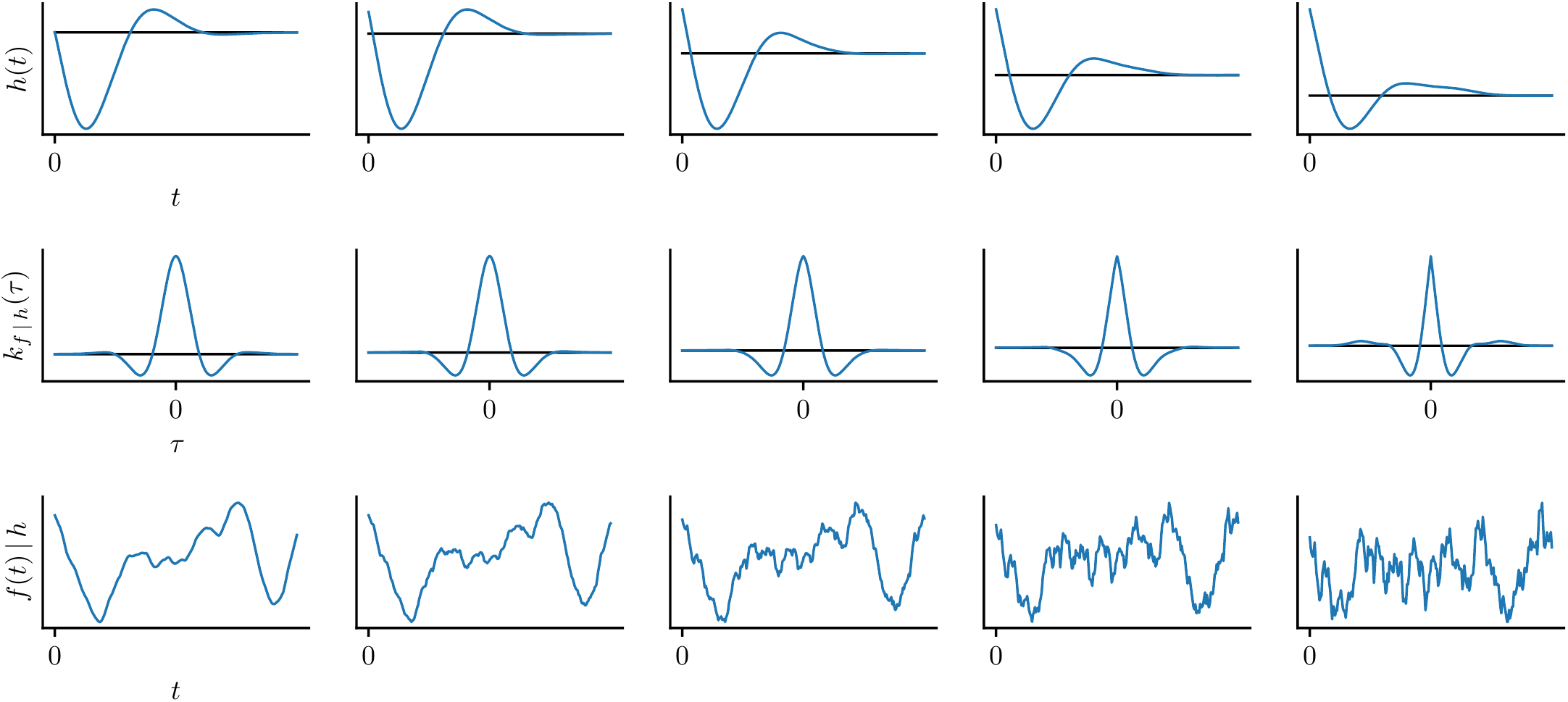}
    \caption{Generative process of the CGPCM. Shows the filter $h$, the kernel $k_{f\cond h}$, and a sample $f\cond h \sim \mathcal{GP}(0,k_{f\cond h}(t-t'))$ while the filter is interpolated from one that satisfies $h(0)=0$ to one that satisfies $|h(0)|>0$ with a fixed random seed.}
    \label{fig:interpolation}
\end{figure*}

\Cref{fig:interpolation} shows the filter $h$, the kernel $k_{f\cond h}$, and a sample $f\cond h \sim \mathcal{GP}(0,k_{f\cond h}(t-t'))$ while the filter is interpolated from one that satisfies $h(0)=0$ to one that satisfies $|h(0)|>0$. Note that the sample appears smooth for $h(0)=0$ and becomes rougher as $|h(0)|$ increases.

\section{Inference}
\label{sec:inference}
Let $y(t)\cond f(t) \sim \mathcal{N}(f(t),\sigma^2)$ for all $t$ be a noisy version of $f$. Given some observations $e=(y(t_1),\ldots,y(t_n))$ of $y$, we wish to compute $p(f,h,x\cond e)$.

We learn $h$ and $x$ through \textit{inducing points} \citep{Titsias:2009:Variational_Learning}, which reduce the infinite-dimensional quantities to a small number of variables:
\begin{align*}
     u&=(h(t_{u,1}),\ldots,h(t_{u,n_u}))\sim \mathcal{N}(0,K_u), \text{ and}\\
     z&=(s(t_{z,1}),\ldots,s(t_{z,n_z}))\sim \mathcal{N}(0,K_z)
\end{align*}
for respectively the processes $h$ and $s=T[x]$ where
\begin{align*}
    T[x](t)=\int r(t- \tau)x(\tau) \sd{\tau}
\end{align*}
is some interdomain transformation of the white-noise process with filter $r$ \citep{Lazaro-Gredilla:2009:Inter-Domain_Gaussian_Processes_for_Sparse,Alvarez:2010:Efficient_Multioutput_Gaussian_Processes_Through,Tobar:2015:Learning_Stationary}. The inter-domain transformation is necessary because white noise is uncorrelated, meaning that placing inducing points directly in this domain would not capture meaningful structure in the posterior. In choosing $r$, we make sure that $s=T[x]$ has power at the majority of frequencies present in the signal that we aim to model, thereby enabling the inducing points to capture posterior dependencies in the components of the white noise that pass through the filter and therefore affect the data. We follow \citet{Tobar:2015:Learning_Stationary} and let $r(t)=\exp(-\omega t^2)$, which fits the low frequency components of the posterior.

Let the \textit{mean-field} (MF) approximation
\begin{align*}
    q(f,u,z)=p(f\cond u, z)q(u)q(z)\approx p(f,u,z\cond e)
\end{align*}
be such that it is closest in Kullback-Leibler divergence. Contrary to the formulation by \citet{Tobar:2015:Learning_Stationary}, we have analytically integrated out $h$ and $x$ in $p(f\cond u, z)=\int p(f, h, x\cond u, z)\sd{h}\sd{x}$ prior to performing inference. It is indeed the case that $p(f\cond u, z)$ is intractable, but its first two moments can in fact be computed, which is sufficient to formulate our inference scheme.
Denote expectation with angle brackets: $\la f(x) \ra_{p(x)} = \int f(x) p(x) \sd{x}$.
Rearranging
\begin{align*}
    &\log p(e) - \KL(q(f,u,z)\,\|\,p(f,u,z\cond e)) \\
    &\quad= \ll\la \log \frac{p(e\cond f)\cancel{p(f\cond u, z)}p(u)p(z)}{\cancel{p(f\cond u, z)}q(u)q(z)} \rr\ra_{q(f,u,z)} \\
    &\quad= \underbrace{\la \log p(e\cond f) \ra_{q(f)}}_{\text{reconstruction cost}} \\
    &\quad\phan{=}- \underbrace{\ll(\ll\la\log\frac{q(u)}{p(u)}\rr\ra_{q(u)} + \ll\la\log\frac{q(z)}{p(z)}\rr\ra_{q(z)}\rr)}_{\text{divergence from prior}} \\
    &\quad=\mathcal{L}[q(u),q(z)]
\end{align*}
shows that we can find $q(u)$ and $q(z)$ through maximising $\mathcal{L}$. Since $\log p(e)\ge\mathcal{L}$, $\mathcal{L}$ is called the \textit{evidence lower bound} (ELBO). Observe that maximising $\mathcal{L}$ attempts to explain the data well whilst not diverging too far from the model prior.

To optimise $\mathcal{L}$ with respect to $q(u)$ and $q(z)$ we set their respective variations $\delta \mathcal{L} / \delta q(u)$ and $\delta \mathcal{L} / \delta q(z)$ to zero; then solving for $q(u)$ and $q(z)$ yields
\begin{align}
    q(u) &\propto p(u) \exp \la \log p(e\cond f) \ra_{p(f\cond u,z)q(z)}, \label{eq:qu} \\
    q(z) &\propto p(z) \exp \la \log p(e\cond f) \ra_{p(f\cond u,z)q(u)}, \label{eq:qz}
\end{align}
which are computed in \cref{app:computation_quqz}.
Note that \cref{eq:qu,eq:qz} are coupled.
Section B of the supplementary material shows that the mean and variance of $p(f\cond u, z)$ are respectively linear and quadratic in both $u$ and $z$; therefore, since $p(e\cond f)$ is Gaussian, we can find a stationary point of $\mathcal{L}$ in which both $q(u)$ and $q(z)$ are Gaussian. Hence, to find $q(u)$ and $q(z)$, we can initialise $q(u)$ and $q(z)$ to some arbitrary Gaussian and either iterate \cref{eq:qu,eq:qz} or maximise $\mathcal{L}$ directly using gradient-based optimisation. In the latter case we can include any hyperparameters in the optimisation too.

One of the new contributions of this work is the finding that we can significantly speed up the optimisation by solving for either $q(u)$ or $q(z)$ analytically. Substituting the optimal form of $q(z)$ back into $\mathcal{L}$ yields
\begin{align}
    \mathcal{L}^*[q(u)]&=\max_{q(z)} \mathcal{L}[q(u),q(z)]\nonumber\\
    &= \log \int p(z)\exp \la \log p(e\cond f) \ra_{p(f\cond u, z)q(u)} \sd{z} \nonumber\\
    &\phan{=}-\ll\la\log\frac{q(u)}{p(u)}\rr\ra_{q(u)},\label{eq:saturated_elbo}
\end{align}
which is computed in \cref{app:saturated_elbo}. We optimise this saturated lower bound $\mathcal{L}^*$ to yield $q(u)$ and then obtain $q(z)$ through \cref{eq:qz}.

Variational mean-field approaches to inference, like the one just presented, are often computationally efficient, but they are known to suffer from certain biases \citep{MacKay:2002:Information_Theory_Learning,Turner:2011:Two_Problems_With_Variational_Expectation,Murphy:2012:Probabilistic_Perspective}. We further refine the MF approximation to alleviate these biases, which forms the second major improvement to inference and learning provided by this work.

To this end, let the \textit{structured mean-field} (SMF) approximation
\begin{align*}
    q(f,u,z)=p(f\cond u, z)q(u,z)\approx p(f,u,z\cond e)
\end{align*}
be such that it again is closest in Kullback-Leibler divergence.
We let $q(u,z)$ be free form, which means that the only assumption underlying the SMF approximation is sufficiency of $u$ and $z$ for respectively $h$ and $s$.
Hence, the SMF approximation is guaranteed be close to the true posterior if there are sufficiently many $u$ and $z$.

Following a similar argument to the above, we again derive the corresponding ELBO:
\begin{align*}
    &\mathcal{L}[q(u),q(z\cond u)] \\
    &\quad= \la \log p(e\cond f) \ra_{q(f)}- \ll\la\log\frac{q(u)q(z\cond u)}{p(u)p(z)}\rr\ra_{q(u)q(z\cond u)}.
\end{align*}
Again, setting the variations $\delta \mathcal{L} / \delta q(u)$ and $\delta \mathcal{L} / \delta q(z\cond u)$ to zero and solving for respectively $q(u)$ and $q(z\cond u)$ yields
\begin{align}
    q(u) &\propto p(u) \int p(z) \exp\la\log p(e\cond f)\ra_{p(f\cond u, z)}\sd{z}, \label{eq:qu-smf} \\
    q(z\cond u) &\propto p(z)\exp\la \log p(e\cond f)\ra_{p(f\cond u, z)}, \label{eq:qz-smf}
\end{align}
which are computed in \cref{app:computation_quz}.
As opposed to \cref{eq:qu,eq:qz}, \cref{eq:qu-smf,eq:qz-smf} are uncoupled, but $q(u)$'s moments are now intractable. We can, however, evaluate the right-hand side of \cref{eq:qu}, which is proportional to $q(u)$. Noting that $p(u)$ is Gaussian, we employ elliptical slice sampling (ESS) \citep{Murray:2010:Elliptical_Slice_Sampling} to sample from $q(u)$ and use these samples to approximate $q(f, u, z)$. To help mixing the Markov chain, we initialise the sampler with the MF approximation.

\section{Experiments}
We evaluate the CGPCM by applying it to synthetic and real-world signals. We show that in certain situations the causality constraint provides an inductive bias that leads to better predictions. Initialisation of the hyperparameters for both the GPCM and CGPCM is discussed in \cref{app:hyperparameters}.

\subsection{Learning Performance}
\begin{figure}[t]
    \centering
    \includegraphics[width=\linewidth]{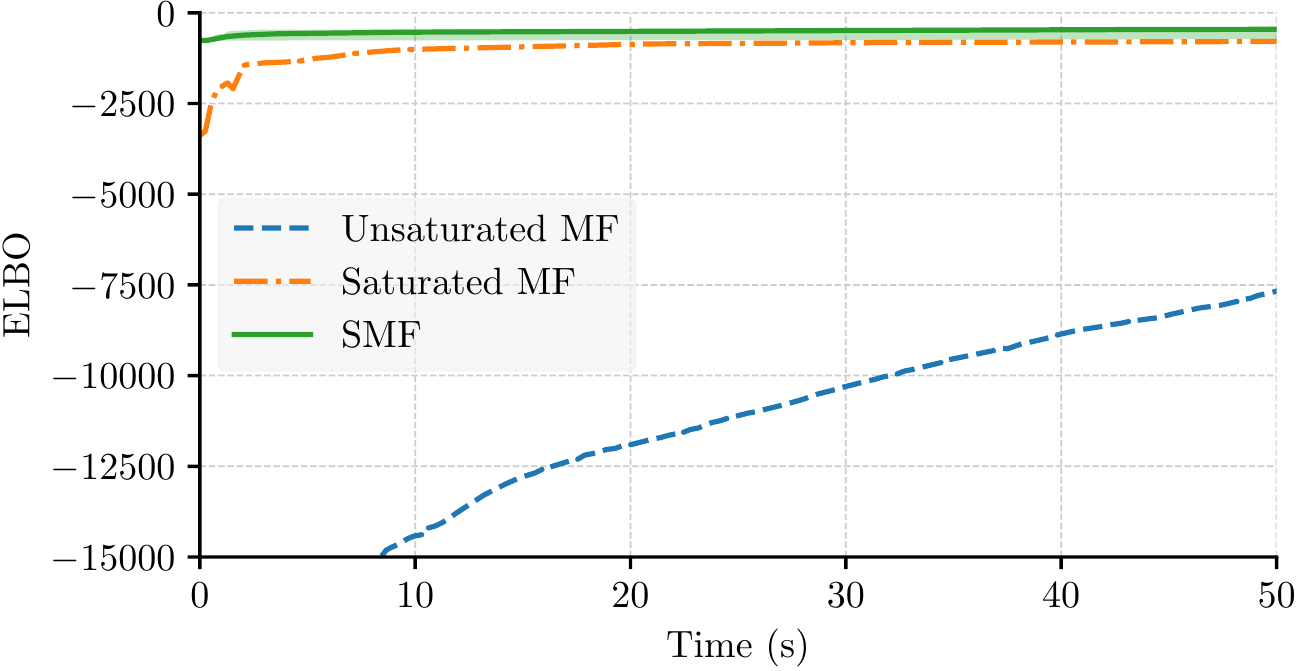}
    \caption{Learning 600 noisy data points sampled from a Gaussian process with a kernel that is a sum of multiple exponentiated-quadratic kernels with various length scales. Shows the evolution of ELBO in time as the unsaturated and saturated MF ELBO are optimised, and the mean and confidence intervals of an Monte Carlo estimate of the ELBO whilst sampling from the SMF approximation. The processes $x$ and $s$ were induced on respectively 150 and 101 points. All parameters were initialised randomly, and \texttt{scipy}'s implementation of the L-BFGS-B algorithm \citep{Nocedal:2006:Numerical_Optimisation} was used to optimise the unsaturated and saturated MF ELBO.}
    \label{fig:opt}
\end{figure}

We compare the presented saturated MF ELBO and the SMF approximation to the original inference scheme by \citet{Tobar:2015:Learning_Stationary}, which we call the \textit{unsaturated} MF ELBO, by comparing learning curves.
\Cref{fig:opt} shows the evolution of the ELBO in time for the three inference schemes as a noisy sample from a GP with a complicated kernel is learned. Observe that the saturated MF ELBO converges quicker and attains a higher value than the unsaturated MF ELBO. Further observe that the SMF approximation nearly immediately converges and attains an even higher value than the saturated MF ELBO; this shows that it is indeed beneficial to model correlation between $u$ and $z$ in the approximate posterior.

\subsection{Synthetic Data Comparison}
\label{subsec:toy}
In \cref{subsec:roughness} we showed that the CGPCM models both smooth and rough signals, whereas the GPCM models only smooth signals. We perform two experiments that show the significance of this modelling capability.

\begin{figure*}[t]
    \centering
    \includegraphics[width=\linewidth]{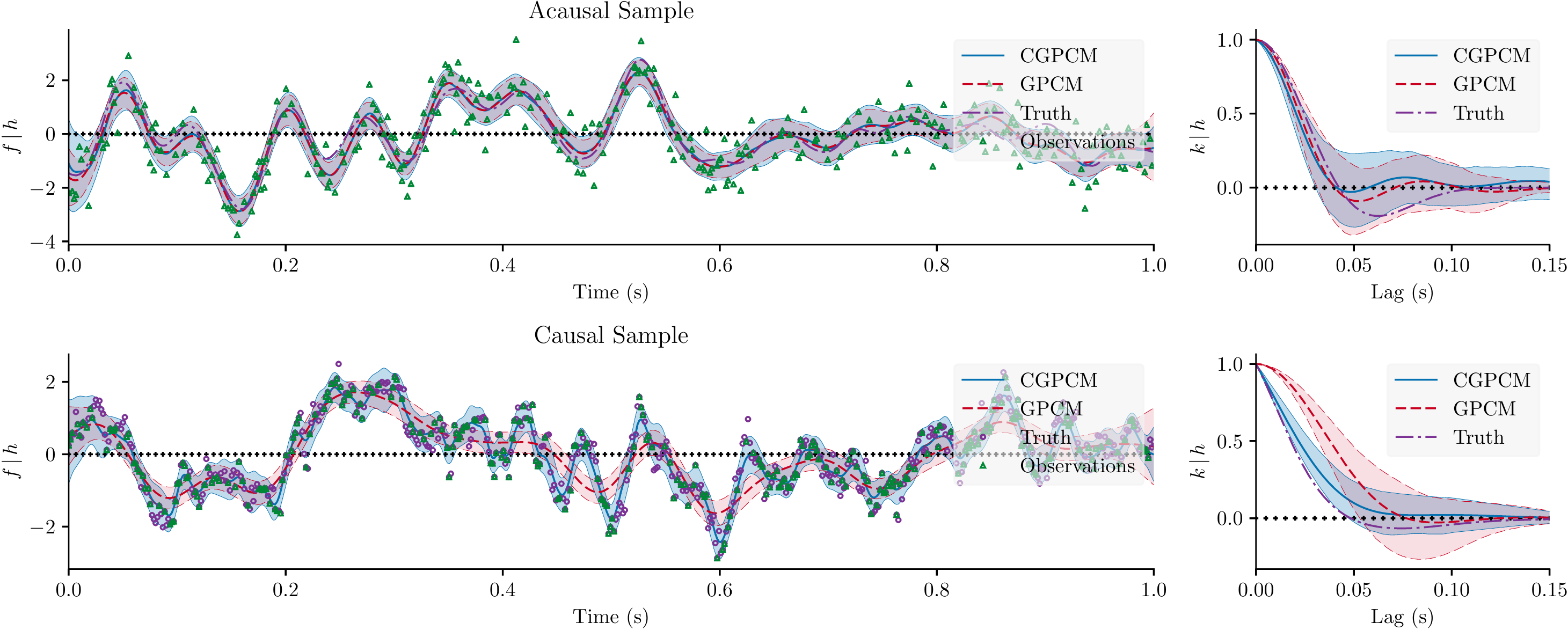}
    \caption{Learning a randomly chosen 200 out of 400 data points from a noisy sample of the GPCM (an acausal sample) and from a noiseless sample from the CGPCM (a causal sample). Furthermore, $x$ and $s$ were induced on respectively 150 and 51 points; the locations of the inducing points are indicated by plusses. All parameters were initialised randomly, and the procedure from \cref{sec:inference} was followed. Finally, all signals were normalised to unity power.}
    \label{fig:toy}
\end{figure*}

\Cref{fig:toy} shows the results of fitting the GPCM and CGPCM to a noisy random sample from the GPCM: an acausal sample. Observe that the GPCM and CGPCM both succesfully fit the function $f \cond h$ and the kernel $k_{f\cond h}$. This confirms that both models are able to succesfully nonparametrically learn an unknown covariance function.

\Cref{fig:toy} further shows the results for fitting the GPCM and CGPCM to instead a \textit{noiseless} random sample from the CGPCM: a causal sample. The challenge for the models is to not mistake the sample roughness for observation noise. Observe that the CGPCM fits the kernel $k_{f \cond h}$ well, correctly capturing the sharpness of $k_{f \cond h}$ at the origin. On the other hand, despite approximately capturing the correlation structure, the GPCM mistakes the sample roughness for observation noise and poorly fits of the function $f\cond h$; this is further reflected in the fit of the kernel $k_{f\cond h}$, which shows that the GPCM is unable to capture the sharpness of $k_{f \cond h}$ at the origin. Note that the latter observation is consistent with \cref{fig:prior_kernel}, which suggests that only the CGPCM gives support to kernels that are nondifferentiable at the origin.

The key difference between the acausal and causal sample in \cref{fig:toy} is that in the former noise is added independently, whereas in the latter noise is part of the process dynamics. Hence, \cref{fig:toy} indicates that both models can learn smooth processes with independently added noise, but only the CGPCM was able to succesfully infer noisy dynamics.

\subsection{Crude Oil Prices Regression}

\begin{figure*}[t]
    \centering
    \includegraphics[width=\linewidth]{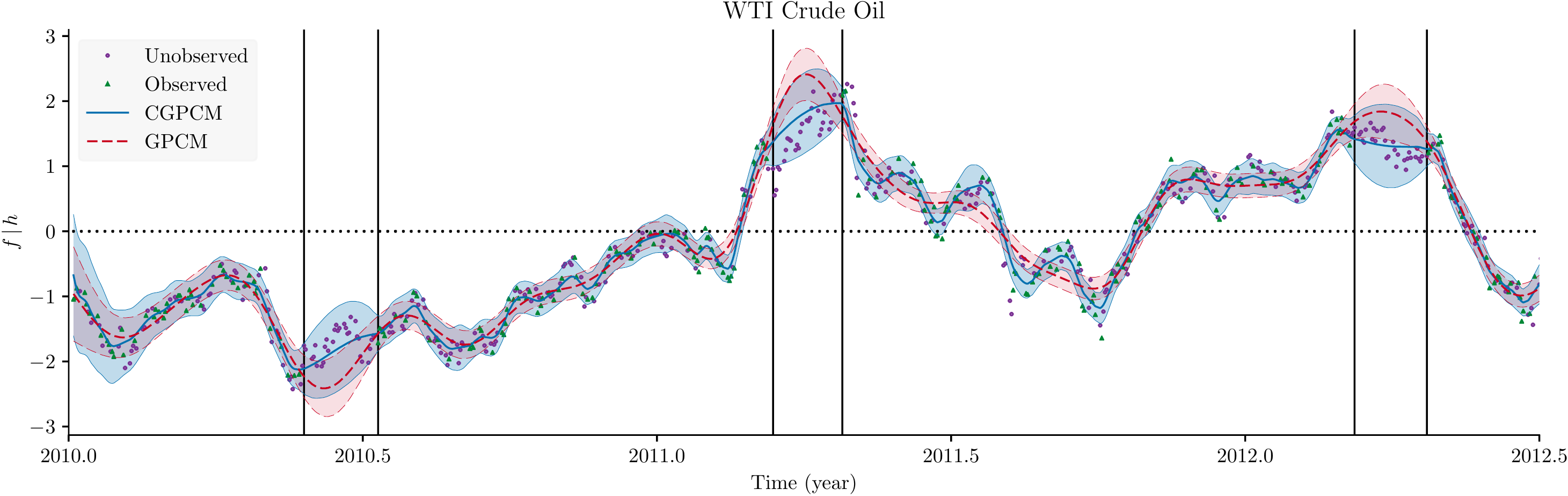}
    \caption{Learning a randomly chosen 400 data points out of daily crude oil prices (TWI crude oil; obtained from \url{http://www.tradingeconomics.com/commodity/crude-oil}) from 2010 to 2014. For clarity, only up to halfway 2012 is shown. Five randomly chosen sections (whereof three are depicted) of 30 data points were omitted to assess the algorithms's ability to extrapolate. The processes $x$ and $s$ were induced on respectively 200 and 51 points; the locations of the inducing points are indicated by plusses. All parameters were initialised randomly, and the procedure from \cref{sec:inference} was followed. Finally, all signals were normalised to unity power.}
    \label{fig:crude}
\end{figure*}

We evaluate the GPCM's and CGPCM's predictive performance by interpolating and extrapolating daily crude oil prices from 2010 to 2014. \Cref{fig:crude} visualises a part of the data set and overlays the models's predictions. Observe that the GPCM and CGPCM provide qualitative different predictions: the CGPCM models the entire variability of the data, whereas the GPCM attributes a part of this variability to noise, hence providing a more smooth prediction (c.f.\ \cref{subsec:toy}). As a consequence, the credible region of the GPCM's prediction fails to capture some unobserved points that lie within the data set. Additionally, the smoothness of the GPCM's prediction causes the prediction to overshoot in regions outside of the data set.

\Cref{tab:stats_crude} shows that the CGPCM outperforms the GPCM in both the tasks of interpolating and extrapolating the daily crude oil prices, and shows that the SMF approximation consistently yields a higher ELBO, but does not yield marked improvements in terms of MLL and SMSE.
\Cref{tab:stats_crude} further shows the results for conventional GP regression with an exponentiated quadratic (EQ) kernel, rational quadratic (RQ) kernel, and Matern-$\nu$ (M$\nu$) kernels \citep{Rasmussen:2006:Gaussian_Processes}.
For the interpolation task, the SMSE achieved by the GPs ranges from worse than the GPCM (EQ) to on par with the CGPCM (RQ and Matern-\sfrac{1}{2}); in all cases, the MLL is better than the GPCM, and only the RQ and Matern-\sfrac{1}{2} kernel come close to the CGPCM.
For the extrapolation task, the SMSE and MLL achieved by the GPs are in all cases better than the GPCM, and again only the RQ comes close to and the Matern-\sfrac{1}{2} kernel even improves over the CGPCM.

Furthermore, note that the likelihoods achieved by conventional GP regression greatly exceed the ELBOs achieved by the CGPCM and GPCM, which can partly be attributed to looseness of the ELBOs.
When comparing the (C)GPCM to conventional GPs, it is important to consider that a sufficiently rich set of parametric covariance functions will contain for each task a kernel with suitable prior assumptions, meaning that after hyperparameter optimisation the GP will perform well.
An advantage of the (C)GPCM is that it automatically infers its kernel, so there is no process of finding a suitable kernel; consequently, the (C)GPCM will perform well across all tasks, though possibly not best.
In automatically inferring its kernel, the (C)GPCM averages over all possible kernels, and the average likelihood resulting from averaging over these unfortunately mostly unsuitable kernels can be expected to be lower than that resulting from hyperparameter optimisation in a conventional GP with a suitable kernel.

In conclusion, the automatic kernel selection performed by the CGPCM and GPCM is for this particular task only rewarding for CGPCM.
The GPCM tends to attribute a part of the variability of the data to noise, performing worse than conventional GPs.
The inductive bias provided by CGPCM, on the other hand, allows the entire variability of the data to be modelled, yielding performance comparable to the best-performing kernels in conventional GPs.
Finally, note that RQ kernel, which gives rise to correlations on multiple length scales, and the Matern--\sfrac{1}{2} kernel, which decays quickly at the origin, tend to perform best: it is exactly this flexibility---complicated covariance functions that may decay quickly at the origin---that the CGPCM offers (cf.\ \cref{fig:interpolation}).

\subsection{Head-Related Impulse Response Estimation}
The head-related impulse response (HRIR) describes how sound entering the ear canal is filtered by the outer ear. The form of the filter depends on the location of the sound source. The filter introduces complex spectral cues into the sounds that are used by the brain to infer source location. We consider the problem of inferring the HRIR from an incoming sound pressure waveform. Suppose that the sound is background noise and can thus be modelled by white noise. Then, assuming a Gaussian process prior over the HRIR, we recover the (C)GPCM exactly; in other words, we can infer the HRIR via inferring the filter $h$ in the (C)GPCM.

\begin{figure*}[t]
    \centering
    \includegraphics[width=\linewidth]{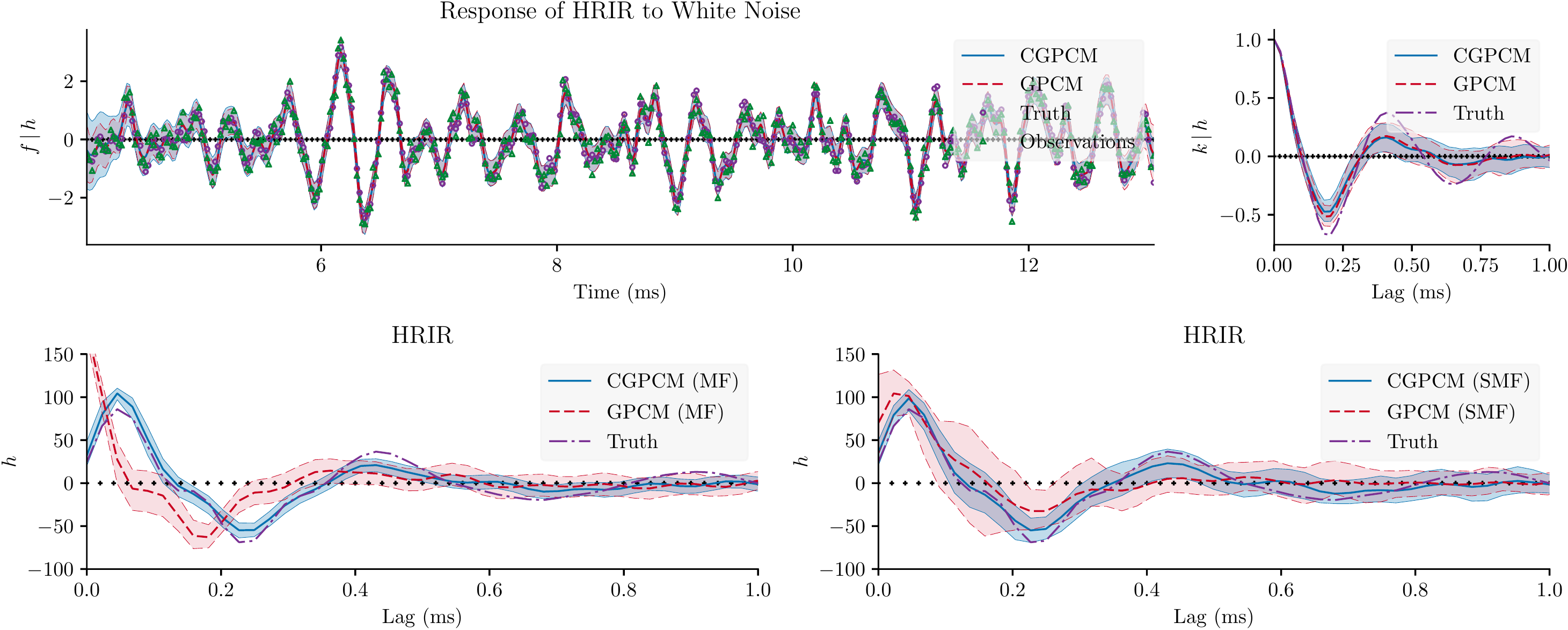}
    \caption{Learning 400 noisy data points of white noise passed through a HRIR downloaded from KEMAR---see \url{http://sound.media.mit.edu/resources/KEMAR.html}. The processes $x$ and $s$ were induced on respectively 200 and 151 points; the locations of the inducing points are indicated by plusses. All parameters were initialised randomly, and the procedure from \cref{sec:inference} was followed. Finally, all signals were normalised to unity power.}
    \label{fig:hrir}
\end{figure*}

We convolve a HRIR from a KEMAR dummy head microphone with white noise to simulate the signal that would have been sensed by the ear. \Cref{fig:hrir,tab:stats_hrir} show the results of fitting the GPCM and CGPCM to the resulting signal. Despite the fact that both models correctly infer the kernel $k_{f \cond h}$, only the CGPCM is able to provide a good prediction for the HRIR. This shows that the inductive bias built in by the causality constraint can help when dealing with causal phenomena. \Cref{tab:stats_hrir} further shows that the SMF approximation consistently yields a higher ELBO, and additionally yields significant improvements in terms of MLL and SMSE.

\section{Discussion}
We have presented the CGPCM as a model of causal, spectrally complex dynamical phenomena. Inference in the (C)GPCM is performed in a two-step procedure: first a collapsed variational MF approach is used to obtain an initial approximation, and then this approximation is refined through a variational SMF approach combined with ESS. The proposed enhanced inference schemes and structered approximating distributions have been shown to greatly improve upon the original inference scheme by \citet{Tobar:2015:Learning_Stationary}. The CGPCM has further been tested on synthesised and real-world signals and shows encouraging results. In particular, the CGPCM shows the capability of modelling rough signals without attributing this roughness to observation noise.

Future research can be taken into various directions. First, the model can be extented to multi-dimensional input and output spaces by following a construction similar to that by \citet{Bruinsma:2016:GGPCM}. Second, the application of more efficient sparse approximation for $h$ and $x$, such as the tree-structured approximation by \citet{Bui:2014:Tree-Structured_Gaussian}, could scale the (C)GPCM to larger data sets. Third, to deal with signals that are band limited, but not necessarily baseband, the use of an harmonic interdomain transformation \citep{Tobar:2015:Inter-Domain_Inducing} could be explored.
Finally, using an approach similar to that by \citet{Lazaro-Gredilla:2013:Variational_Inference_for_Mahalanobis_Distance}, Bayesian inference in the hyperparameters could be attempted.

\clearpage
\begin{table*}[t]
    \centering
    \caption{Statistics accompanying the experiment performed in \cref{fig:crude}. Also shows the results for conventional GP regression with an exponentiated-quadratic (EQ) kernel, rational quadratic (RQ) kernel, and various Matern-$\nu$ (M$\nu$) kernels. SMSE refers to the standardised mean squared error, MLL refers to the mean log loss, and LIK.\ refers to the model evidence. \citep{Rasmussen:2006:Gaussian_Processes}}
    \label{tab:stats_crude}
    \vskip 0.15in
    \centering
    \begin{small}
        \begin{sc}
            \begin{tabular}{lccc}
                \toprule
                \multicolumn{4}{c}{WTI Crude Oil Interpolation} \\ \midrule
                & SMSE & MLL & LIK. \\ \midrule
                GP (EQ) & \SI{2.90e-1} & \SI{9.69e-1} & \SI{1.79} ~\\
                GP (RQ) & \SI{4.91e-2} & \SI{2.46e-1} & \SI{3.94e1} ~\\
                GP (M\sfrac{1}{2}) & \SI{5.18e-2} & \SI{2.30e-1} & \SI{4.91e1} ~\\
                GP (M\sfrac{3}{2}) & \SI{1.05e-1} & \SI{5.33e-1} & \SI{3.20e1} ~\\
                GP (M\sfrac{5}{2}) & \SI{1.30e-1} & \SI{6.18e-1} & \SI{2.10e1} ~\\ \midrule
                & SMSE & MLL & ELBO \\ \midrule
                GPCM &&& \\
                \quad MF & \SI{1.73e-1} & \SI{3.50} & \SI{-1.63e2} ~\\
                \quad SMF & \SI{1.87e-1} & \SI{3.42} & \SI{-1.56e2} ~\\
                CGPCM \\
                \quad MF & \SI{4.81e-2} & \SI{1.55e-1} & \SI{-1.46e2} ~\\
                \quad SMF & \SI{5.33e-2} & \SI{2.14e-1} & \SI{-1.36e2} ~\\
                \bottomrule
            \end{tabular}
        \end{sc}
    \end{small}
    \begin{small}
        \begin{sc}
            \begin{tabular}{lccc}
                \toprule
                \multicolumn{4}{c}{WTI Crude Oil Extrapolation} \\ \midrule
                & SMSE & MLL & LIK. \\ \midrule
                GP (EQ) & \SI{4.04e-2} & \SI{9.16e-1} & \SI{1.79} ~\\
                GP (RQ) & \SI{3.28e-2} & \SI{-7.61e-2} & \SI{3.94e1} ~\\
                GP (M\sfrac{1}{2}) & \SI{3.13e-2} & \SI{-4.15e-1} & \SI{4.91e1} ~\\
                GP (M\sfrac{3}{2}) & \SI{3.42e-2} & \SI{2.06e-2} & \SI{3.20e1} ~\\
                GP (M\sfrac{5}{2}) & \SI{3.61e-2} & \SI{5.02e-1} & \SI{2.10e1} ~\\ \midrule
                & SMSE & MLL & ELBO \\ \midrule
                GPCM \\
                \quad MF & \SI{6.21e-2} & \SI{1.42} & \SI{-1.63e2} ~\\
                \quad SMF & \SI{6.08e-2} & \SI{1.28} & \SI{-1.56e2} ~\\
                CGPCM \\
                \quad MF & \SI{3.60e-2} & \SI{-2.32e-1} & \SI{-1.46e2} ~\\
                \quad SMF & \SI{3.62e-2} & \SI{-2.29e-1}  & \SI{-1.36e2} ~\\
                \bottomrule
            \end{tabular}
        \end{sc}
    \end{small}
    \vskip -0.1in
\end{table*}

\clearpage
\begin{table*}[t]
    \centering
    \caption{Statistics accompanying the experiment performed in \cref{fig:hrir}. SMSE refers to the standardised mean squared error, and MLL refers to the mean log loss. \citep{Rasmussen:2006:Gaussian_Processes}}
    \label{tab:stats_hrir}
    \vskip 0.15in
    \centering
    \begin{small}
        \begin{sc}
            \begin{tabular}{lccc}
                \toprule
                \multicolumn{4}{c}{HRIR Estimation} \\ \midrule
                & SMSE & MLL & ELBO \\
                GPCM \\
                \quad MF & \SI{1.33e0} & \SI{1.11e1} & \SI{-4.04e2} ~\\
                \quad SMF & \SI{3.93e-1} & \SI{4.36e0} & \SI{-3.88e2} ~\\
                CGPCM \\
                \quad MF & \SI{1.22e-1} & \SI{4.73e0} & \SI{-4.07e2} ~\\
                \quad SMF & \SI{1.03e-1} & \SI{3.57e0} & \SI{-3.97e2} ~\\
                \bottomrule
            \end{tabular}
        \end{sc}
    \end{small}
    \vskip -0.1in
\end{table*}

\clearpage
\section*{Acknowledgements}
Richard E. Turner is supported by Google as well as EPSRC grants EP/M0269571 and EP/L000776/1.

\bibliography{bibliography}
\bibliographystyle{icml2018}

\clearpage
\appendix

\section{Initialisation of Hyperparameters}
\label{app:hyperparameters}
Following \citet{Pavliotis:2012:Probabilistic_Perspective}, let the \textit{correlation time} $\tau$, or alternatively the \textit{length scale}, of a stationary process with kernel $k(r)$ be
\begin{align*}
    \tau = \frac{1}{k(0)}\int_0 k(r) \sd{r}.
\end{align*}

Let $\tau\ss{ac}$ and $P\ss{ac}$ and $\tau\ss{c}$ and $P\ss{c}$ be respectively the correlation time and power of the process $f$ for the GPCM and CGPCM respectively. Then straightforward calculation shows that $\tau_w=\sqrt{\pi/8 \alpha}$ and
\begin{align*}
    P\ss{ac} &= \sigma_{f,\text{ac}}^2\sqrt{\frac{\pi}{2\alpha\ss{ac}}},
    &\tau\ss{ac} &= \sqrt{\frac{\pi}{2(\alpha\ss{ac} + 2 \gamma\ss{ac})}}  \\
    P\ss{c} &= \sigma_{f,\text{c}}^2\sqrt{\frac{\pi}{8\alpha\ss{c}}},
    &\tau\ss{c} &= \sqrt{\frac{2}{\pi(\alpha\ss{c} + 2 \gamma\ss{c})}} \\
    &&&\phan{=}\atan\ll( \sqrt{\frac{\alpha\ss{c} + 2 \gamma\ss{c}}{\alpha\ss{c}}} \rr). \\
\end{align*}
It follows that $P\ss{c}=\frac{1}{2}P\ss{ac}$ and $\tau\ss{c}\approx \tau\ss{ac}$ if $\gamma\ss{c} \gg \alpha\ss{c}$, $\alpha\ss{c}=\alpha\ss{ac}$, $\gamma\ss{c}=\gamma\ss{ac}$, and $\sigma_{f,\text{c}}^2=\sigma_{f,\text{ac}}^2$. To allow for fair comparison, we require that $P\ss{ac}=P\ss{c}$ and $\tau\ss{ac}=\tau\ss{c}$. If we further choose $P\ss{ac}=P\ss{c}=1$ and $\sigma_{f,\text{ac}}^2=\sigma_{f,\text{c}}^2$, then we find that $\alpha\ss{c}=\frac{1}{4}\alpha\ss{ac}$, $\gamma\ss{c}=\frac{3}{8}\alpha\ss{ac}+ \gamma\ss{ac}$, and $\sigma_{f,\text{ac}}^2 = \sigma_{f,\text{c}}^2 = \sqrt{2\alpha\ss{ac}/\pi}$.

To determine $\alpha\ss{ac}$ and $\gamma\ss{ac}$, we specify the correlation time $\tau_w$ of the window $w$ and the correlation time $\tau_f$ of the process $f$. Then solving yields that
\begin{align*}
    \alpha\ss{ac} &= \frac{\pi}{2} \cdot \frac{1}{2 \tau_w^2},
    &\gamma\ss{ac} &= \frac{\pi}{2} \cdot \frac{1}{2 \tau^2_f} - \frac{1}{2} \alpha\ss{ac}.
\end{align*}
Since $\alpha\ss{ac} > 0$ and $\gamma\ss{ac} > 0$, it must hold that $\sqrt{2} \tau_w > \tau_f > 0$.

Furthermore, for the GPCM, we let $t_u$ be evenly spaced from $-3\tau_w$ to $3 \tau_w$. For the CGPCM, we let $t_u$ be evenly spaced from zero to $6 \tau_w$, shifted two inter-point spacings $\Delta t_u$ to the left to account for derivatives of $h$ at zero.

Finally, we let $t_z$ be evenly spaced on the domain of interest, and choose $\omega$ such that $s$'s correlation time $\tau_s$ equals twice $t_z$'s inter-point spacing $\Delta t_z$: $\omega = \pi / 8 \Delta t_z^2$.

\section{Moments of $f\cond u, h$}
\label{app:moments_f}
We solve for $\la f(t) \ra_{p(f\cond u,z)}$ and $\la f(t) f(t') \ra_{p(f\cond u,z)}$. First, we have that
\begin{align*}
    &\la f(t) \ra_{p(f\cond u,z)} \\
    &\quad= \int^t \la h(t - \tau)\ra_{p(f\cond u)} \la x(\tau) \ra_{p(x\cond z)} \sd{\tau} \\
    &\quad= u^{\T} K_u^{-1} \underbrace{\int^t k_h(t_u,t-\tau) k_{xs}(\tau, t_z^{\T}) \sd{ \tau}}_{A^{hx}(t)=A^{(xh)\T}(t)} K_z^{-1} z.
\end{align*}
Second, it holds that

\begin{align*}
    &\la f(t) f(t') \ra_{p(f\cond u,z)} \\
    &\quad= \int^t\!\!\!\!\int^{t'} \la h(t- \tau) h(t' - \tau')\ra_{p(h\cond u)} \\
    &\quad\phan{=\int^t\!\!\!\!\int^{t'}} \la x(\tau) x(\tau') \ra_{p(x\cond z)}\sd{\tau'}\sd{\tau} \\
    &\quad= \int^t\!\!\!\!\int^{t'} ( k_h(t- \tau,t' - \tau') \\
    &\quad\phan{=\int^t\!\!\!\!\int^{t'} (}+ k_h(t- \tau, t_u^{\T}) M^u k_h(t_u, t' - \tau')) \\
    &\quad\phan{=\int^t\!\!\!\!\int^{t'}}( k_x(\tau,\tau') + k_{xs}(\tau, t_z^{\T}) M^z k_{sx}(t_z, \tau')) \sd{\tau'}\sd{\tau} \\
    &\quad= a(t,t') + \tr M^u A^{h}(t,t') + \tr M^z A^{x}(t,t') \\
    &\quad\phan{=} + \tr M^u A^{hx}(t) M^z A^{xh}(t')
\end{align*}
where $M^u=K_u^{-1}uu^{\T} K_u^{-1}-K_u^{-1}$, $M^z=K_z^{-1}zz^{\T}K_z^{-1}-K_z^{-1}$, and
\begin{align*}
    a(t,t')&=\int^t\!\!\!\!\int^{t'} k_h(t- \tau,t' - \tau') k_x(\tau,\tau') \sd{\tau'}\sd{\tau} \\
    &=\int^{t \land t'} k_h(t- \tau,t' - \tau)\sd{\tau}, \\
    A^{h}(t,t')&=\int^t\!\!\!\!\int^{t'} k_h(t_{u}, t' - \tau') k_x(\tau,\tau') \\
    &\phan{=\int^t\!\!\!\!\int^{t'}}\, k_h(t- \tau, t_{u}^{\T}) \sd{\tau'}\sd{\tau}\\
    &=\int^{t \land t'} k_h(t_{u}, t' - \tau)k_h(t- \tau, t_{u}^{\T}) \sd{\tau}  \\
    A^{x}(t,t')&=\int^t\!\!\!\!\int^{t'} k_{sx}(t_z, \tau') k_h(t- \tau,t'-\tau') \\
    &\phan{=\int^t\!\!\!\!\int^{t'}}\, k_{xs}(\tau, t_z^{\T}) \sd{\tau'}\sd{\tau}.
\end{align*}
Rearranging, we arrive at $\la f(t) \ra = u^{\T}  A^{hx}(t) z$ and
\begin{align*}
    &\la f(t) f(t') \ra - \la f(t) \ra \la f(t') \ra \\
    &\quad = b(t,t') + u^{\T} B^{h}(t,t') u + z^{\T} B^{x}(t,t') z
\end{align*}
where the expectation is over $p(f\cond K_u^{-1} u, K_z^{-1} z)$ and
\begin{align*}
    b(t,t) &= a(t,t') - \tr K_u^{-1} A^{h}(t,t') - \tr K_z^{-1}A^{x}(t,t') \\
    &\phan{=} + \tr K_u^{-1}A^{hx}(t)K_z^{-1}A^{xh}(t'), \\
    B^{h}(t,t') &= A^{h}(t,t') - A^{hx}(t)K_z^{-1} A^{xh}(t'), \\
    B^{x}(t,t') &= A^{x}(t,t') - A^{xh}(t)K_u^{-1} A^{hx}(t').
\end{align*}
Finally, we denote $a(t)=a(t,t)$ and do so for $A^h$, $A^x$, $A^{hx}$, $b$, $B^h$, and $B^x$ as well.

\section{MF Approximation: Computation of $q(u)$ and $q(z)$}
\label{app:computation_quqz}
Following \cref{app:moments_f}, we have that
\begin{align*}
    &\la \log p(e\cond f) \ra_{p(f\cond K_u^{-1} u, K_z^{-1} z)} \\
    &\quad= -\frac{n}{2}\log 2 \pi \sigma^2 - \frac{1}{2 \sigma^2} \la e^{2}(t) - 2 \sigma_f u^{\T} A^{hx}(t) e(t) z\\
    &\quad\phan{=}  + \sigma_f^2 ( b(t) + u^{\T} B^{h}(t) u+ z^{\T} B^{x}(t) z \\
    &\quad\phan{=}  + (u^{\T} A^{hx}(t) z)^2 ) \ra_t
\end{align*}
where $\la\,\cdot\,\ra_t$ denotes summation with respect to $t$ over $t_1,\ldots,t_n$.
It follows that
\begin{align*}
    &\log p(K_u^{-1} u)  + \la \log p(e\cond f) \ra_{p(f\cond K_z^{-1}z, K_u^{-1}u)q(K_z^{-1}z)} \\
    &\quad= -\frac{1}{2}u^{\T}\ll(K_u + \frac{\sigma_f^2}{\sigma^2} \la B^{h}(t) \rr.\\[-3\jot]
    &\quad\phan{= -\frac{1}{2}u^{\T}}
        \underbrace{
            \phan{\ll(K_u + \frac{\sigma_f^2}{\sigma^2} \la \rr. } \!\!\!\!\!
            \ll. \vphantom{\frac{\sigma_f^2}{\sigma^2} }
                 + A^{hx}(t)  z z^{\T} A^{xh}(t) \ra_{t,q(K_z^{-1} z)}
            \rr)
        }_{\Sigma_u^{-1}}u \\
    &\quad\phan{=}+ u^{\T} \underbrace{\frac{\sigma_f}{\sigma^2}\la e(t) A^{hx}(t) z\ra_{t,q(K_z^{-1}z)}}_{\Sigma_u^{-1} \mu_u} \\
    &\quad\phan{=}  + \ll(\vphantom{\frac{\sigma_f^2}{2 \sigma^2}} {-\frac{n}{2}}\log 2 \pi \sigma^2-\frac{1}{2}\log|2 \pi K_u^{-1}|  \rr. \\
    &\quad\phan{=+}\underbrace{\phan{ \ll(\vphantom{\frac{\sigma_f^2}{2 \sigma^2}}\rr.}- \frac{\la e^2(t)\ra_t}{2 \sigma^2} - \ll.\frac{\sigma_f^2}{2 \sigma^2}\la b(t) - z^{\T} B^{x}(t) z\ra_{t,q(K_z^{-1}z)}\rr)}_{\text{constant independent of $u$}} \\
    &\quad= \log \underbrace{\mathcal{N}(u; \mu_u, \Sigma_u)}_{q(K_u^{-1}u)}+ \frac{1}{2}\log|2 \pi \Sigma_u| \\
    &\quad\phan{=}+ \frac{1}{2}\mu_u^{\T} \Sigma_u^{-1}\mu_u  + \text{constant independent of $u$}
\end{align*}
and $q(z)$ is derived similarly.

\section{MF Approximation: Saturated ELBO}
\label{app:saturated_elbo}
Following \cref{app:computation_quqz}, we have that
\begin{align*}
    &\mathcal{L}^*[q(K_u^{-1} u)] \\
    &\quad = - \frac{n}{2}\log 2 \pi \sigma^2 + \frac{1}{2}\log|K_z^{-1}||\Sigma_z| + \frac{1}{2}\mu_z^{\T} \Sigma_z^{-1}\mu_z  \\
    &\quad \phan{=} - \frac{\la e^2(t)\ra_t}{2 \sigma^2}- \frac{\sigma_f^2}{2 \sigma^2}\la b(t) +  u^{\T} B^{h}(t) u \ra_{t,q(K_u^{-1}u)} \\
    &\quad \phan{=}  + \ll\la \log \frac{p(K_u^{-1}u)}{q(K_u^{-1}u)} \rr\ra_{q(K_u^{-1}u)}.
\end{align*}

\section{SMF Approximation: Computation of $q(u)$ and $q(z\cond u)$}
\label{app:computation_quz}
The mean and variance of $q(z\cond u)$ are similar to that of $q(z)$ in \cref{app:computation_quqz} with only the difference being that the expectation with respect to $q(u)$ is omitted.

To compute $q(u)$, we again follow \cref{app:computation_quqz} and have that
\begin{align*}
    \Sigma &= K_z + \frac{\sigma_f^2}{\sigma^2} \la B^x(t) + A^{xh}(t) u u^{\T}A^{hx}(t) \ra_t, \\
    \mu &= \frac{\sigma_f}{\sigma^2}\la e(t) A^{xh}(t)u\ra_{t}, \\
    \log q(K_u^{-1} u) &= \log p(K_u^{-1} u)- \frac{n}{2}\log 2 \pi \sigma^2 -\frac{1}{2}\log|K_z||\Sigma|\\
    &\phan{=}+ \frac{1}{2}\mu^{\T} \Sigma^{-1} \mu- \frac{\la e^2(t)\ra_t}{2 \sigma^2} \\
    &\phan{=}  - \frac{\sigma_f^2}{2 \sigma^2} \la b(t) + u^{\T} B^h(t)u \ra_t.
\end{align*}

\end{document}